\documentclass{article}

\usepackage{arxiv}
\usepackage[final,nopatch=footnote]{microtype}% Avoid underfull vbox issues
\usepackage[utf8]{inputenc} % allow utf-8 input
\usepackage[T1]{fontenc}    % use 8-bit T1 fonts
\usepackage{hyperref}       % hyperlinks
\usepackage{url}            % simple URL typesetting
\usepackage{booktabs}       % professional-quality tables
\usepackage{amsfonts}       % blackboard math symbols
\usepackage{nicefrac}       % compact symbols for 1/2, etc.
\usepackage{microtype}      % microtypography
\usepackage{lipsum}		% Can be removed after putting your text content
\usepackage{graphicx}
\usepackage[numbers]{natbib}
\usepackage{doi}
\usepackage{amsmath}
\usepackage{graphicx}
\usepackage{amsmath}
\usepackage{orcidlink}
\usepackage{graphicx}
\usepackage{booktabs}
\usepackage{subcaption}
\usepackage{rotating}
\usepackage{makecell}
\usepackage{graphicx}
\usepackage{booktabs}
\usepackage{siunitx}
\usepackage{geometry}
\usepackage{rotate}
\usepackage{longtable}
\usepackage{booktabs}
\usepackage{array}
\usepackage{geometry}
\usepackage{pdflscape}
\usepackage{longtable}
\usepackage{array}
\usepackage{booktabs}
\usepackage{longtable}
\usepackage{booktabs}
\usepackage{tabularx}
\usepackage{float}
\usepackage{setspace}
\title{Enhancing weed detection performance by means of GenAI-based image augmentation \thanks{\textit{This preprint is a revised version of the paper submitted to the CVPPA Workshop at ECCV 2024}}}
\renewcommand\footnotemark{}
\author{%
\begin{minipage}[t]{0.45\textwidth}
    \centering
    {\fontsize{10}{12}\selectfont
    \href{https://orcid.org/0009-0006-7759-0864}{\includegraphics[scale=0.06]{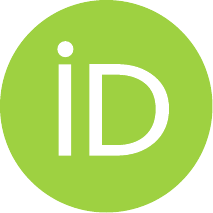}\hspace{1mm} \textbf{Sourav Modak}} \\
    \textnormal{Department of Artificial Intelligence in \\
    Agricultural Engineering \&} \\% Reduce line gap
    \textnormal{Computational Science Hub} \\
    \textnormal{University of Hohenheim} \\
    \textnormal{Garbenstraße 9, 70599 Stuttgart, Germany} \\
    \texttt{s.modak@uni-hohenheim.de} \\
    }
\end{minipage}
\hspace{8mm} % Adjust horizontal gap
\begin{minipage}[t]{0.45\textwidth}
    \centering
    {\fontsize{10}{12}\selectfont
    \href{https://orcid.org/0000-0002-1808-9758}{\includegraphics[scale=0.06]{orcid.pdf}\hspace{1mm} \textbf{Anthony Stein}} \\
    \textnormal{Department of Artificial Intelligence in \\ Agricultural Engineering \&} \\
    \textnormal{Computational Science Hub} \\
    \textnormal{University of Hohenheim} \\
    \textnormal{Garbenstraße 9, 70599 Stuttgart, Germany} \\
    \texttt{anthony.stein@uni-hohenheim.de} \\
    }
\end{minipage}
}

\renewcommand{\shorttitle}{Enhancing Weed Detection Performance through GenAI-Based Image Augmentation}
\begin{document}
\maketitle
%\renewcommand\thefootnote{}
%\footnotetext{\textit{This manuscript is a revised version of the work submitted to the CVPPA workshop at ECCV 2024.
%}}
% Restore footnote numbering for subsequent footnotes
%\renewcommand\thefootnote{\arabic{footnote}}
\begin{abstract}
Precise weed management is essential for sustaining crop productivity and ecological balance. Traditional herbicide applications face economic and environmental challenges, emphasizing the need for intelligent weed control systems powered by deep learning. These systems require vast amounts of high-quality training data. The reality of scarcity of well-annotated training data, however, is often addressed through generating more data using data augmentation. Nevertheless, conventional augmentation techniques such as random flipping, color changes, and blurring lack sufficient fidelity and diversity. This paper investigates a generative AI-based augmentation technique that uses the Stable Diffusion model to produce diverse synthetic images that improve the quantity and quality of training datasets for weed detection models. Moreover, this paper explores the impact of these synthetic images on the performance of real-time detection systems, thus focusing on compact CNN-based models such as YOLO nano for edge devices. The experimental results show substantial improvements in mean Average Precision (mAP50 and mAP50-95) scores for YOLO models trained with generative AI-augmented datasets, demonstrating the promising potential of synthetic data to enhance model robustness and accuracy.
\end{abstract}

\section{Introduction}
\label{sec:intro}
Weed management is pivotal for maintaining productivity and ecological balance in crop production systems, as weeds compete with crops for essential resources such as moisture, sunlight, and nutrients, adversely impacting growth and yield. In scenarios of uncontrolled weed growth, crop yield losses can escalate to 100\%~\cite{kropff1991simple}. The application of herbicides is a common method of weed control; however, the excessive use of chemicals poses significant economic and ecological risks. Site-specific management that uses intelligent weed control systems, such as smart sprays augmented with deep learning-based computer vision, offers a solution to balance crop production with environmental and economic sustainability~\cite{slaughter2008autonomous}. Developing and training deep learning algorithms require large amounts of high-quality data, and data augmentation is a widely adopted technique to mitigate data scarcity.

Traditional data augmentation methods in image processing, including random flipping, color changes, denoising, and blurring, often fall short in fidelity, variation, and natural diversity~\cite{mumuni2022data}. In agricultural contexts, weed infestations exhibit spatiotemporal heterogeneity, and incorporating these variations into synthetic images can significantly enhance the quality of the data set, thereby improving the performance and generalization of weed detection models. In contrast, generative AI-based techniques, such as generative adversarial networks (GANs) and diffusion models, have demonstrated efficacy in preserving fidelity while introducing natural diversity by synthesizing heterogeneous weed representation in the augmented datasets~\cite{modak2024synthesizing}. For example, a recent study~\cite{modak2024synthesizing} demonstrated the efficiency of their advanced text-to-image generation pipeline that uses the Segment Anything Model (SAM)~\cite{kirillov2023segment} in combination with a Stable Diffusion Model~\cite{rombach2022high}. This innovative approach excels in creating highly diverse and realistic datasets, specifically tailored for generating synthetic representations of weed-infested Sugar beet trial plots. However, the work focuses on the generation of highly diverse and realistic synthetic training data, yet not evaluating the impact on downstream task performance. Therefore, we aim to build upon the approach and investigate the impact of generative AI-based image augmentation on the performance of real-time detection systems, such as compact CNN-based models such as YOLO nano. Accordingly, the contributions of our paper are two-fold:

\begin{enumerate}
    \item We investigate the effects of generative AI-based image augmentation by progressively incorporating larger shares of synthetic images into the training dataset of real-time detection systems.
    \item We evaluate and compare the effectiveness of the generative AI-based augmentation approach with conventional image augmentation methods, providing insights into their relative advantages and limitations.
\end{enumerate}

We therefore proceed by providing a brief overview of image augmentation techniques and YOLO models in Section \ref{sec:back} . Our approach is detailed in Section~\ref{sec:meth}. Section \ref{sec:eval} and \ref{sec:dis} describe our reports on the results. Finally, Section \ref{sec:con} concludes with a short discussion on the potential directions of future work.

\section{Background}
\label{sec:back}
%This section demonstrates a brief overview of image augmentation techniques (Sect.\ref{subsec:img_aug}), and the state-of-the-art You Only Look Once (YOLO) models (Sect.\ref{subsec:yolo}).

\subsection{Data Augmentation}
\label{subsec:img_aug}

%Data augmentation is the most appropriate approach to deal with class imbalance, overfitting, and increase diversity in the training dataset to improve the overall training accuracy of the images. 
In general, popular image enhancement techniques are classified as model-free image transformation, model-based synthetic image generation, and hybrid techniques~\cite{nitin2023developing}. Model-free image transformation techniques involve photometric transformations, such as blurring, adding noise, and alterations of the color space, and geometric transformations, such as rotation and scaling~\cite{mumuni2022data}. These model-free techniques can improve the performance of downstream models based on DL. However, the generated images from model-free augmentation are limited to variations of the input images, which cannot explore the images' complete feature space and cannot generate new realistic scenes and detailed information. This limitation can lead to overfitting issues~\cite{divyanth2022image}. However, two main strategies exist: online and offline augmentation. Online augmentation occurs during training, conserving memory but slowing down the process. Offline augmentation pre-generates data, speeding up training but using more memory~\cite{morid2021scoping}.

Contrarily, model-based techniques leverage advanced generative AI models such as \textit{GANs}~\cite{goodfellow2014generativeadversarialnetworks} and \textit{Diffusion Models}~\cite{ho2020denoising} to overcome the constraints of model-free image augmentation methods. Moreover, contemporary generative models can generate images from so-called \textit{text prompt}s, facilitating the synthesis of novel scenes. This capability significantly improves the diversity and robustness of the data set~\cite{zhou2024survey}. Furthermore, \textit{image-to-image translation} can effectively modify the image domain, such as adjusting weather conditions, soil types, and species (e.g., transforming white cabbage to red cabbage and vice versa), thereby significantly enhancing dataset diversity~\cite{luling2024unsupervised}. Furthermore, the approach outlined in~\cite{incollection} demonstrates that synthetic datasets effectively supplement real data in training machine learning models for weed detection.
%This enhancement plays a crucial role in facilitating robust model training. 

\subsection{You Only Look Once (YOLO) models}
\label{subsec:yolo}

There are two types of CNN-based object detectors exist: the so-called \textit{two-stage detectors (region-based)} and \textit{single-stage detectors (regression-based)}. Despite having better accuracy, \textit{two-stage detectors} have been overshadowed in recent years by \textit{single-stage detectors} due to their superior performance in real-time detection. A popular single-stage detector model family is YOLO~\cite{redmon2016you}, which introduced a groundbreaking approach to real-time object detection due to its balance between speed and accuracy~\cite{wang2023comprehensive}. YOLO takes the input image and divides it into a $ S \times S $ grid, where each grid cell predicts bounding boxes $ B $ and class probabilities $ C $. The output bounding boxes consist of the center coordinates $ (bx, by) $, box height $ bh $, width $ bw $, and confidence score $ Pc $. Predictions benefit from \textit{non-max suppression (NMS)} to remove redundant bounding boxes by selecting the box with the highest prediction score and suppressing overlapping boxes~\cite{wang2023comprehensive}. Due to its effectiveness in real-time object detection, YOLO models have evolved rapidly (YOLOv9 and YOLOv10), showing significant improvements over the popular YOLOv8 known for its flexibility and performance. YOLOv9 models addressed challenges from YOLOv8 by introducing the \textit{Programmable Gradient Information (PGI)} technique to preserve information during forward passes. Additionally, the \textit{Generalized Efficient Layer Aggregation Network (GELAN)} improved computational efficiency by replacing depthwise convolutions with conventional convolution operators in the inference step. YOLOv10~\cite{wang2024yolov10}, the current state-of-the-art model in the YOLO family, further improved real-time object detection by enhancing computational efficiency, post-processing efficiency, and model accuracy. Key architectural advancements in YOLOv10 include the introduction of \textit{CSPNet (Cross Stage Partial Network)} for feature extraction and \textit{PAN (Path Aggregation Network)} layers for effective multiscale feature fusion. In particular, YOLOv10 achieves computational efficiency through \textit{NMS-free} training and the \textit{one-to-one head} technique. YOLO models typically come with \textit{nano, small, medium, large}, and \textit{extra large}. When it comes to the deployment on resource-constrained edge computers (e.g., NVIDIA Jetson series\footnote{\url{https://developer.nvidia.com/embedded/jetson-modules}(accessed on: 04 July 2024)}), YOLO nano models come into play for its smaller computational resources, low latency, and high accuracy in real-time object detection. The use of nano-models has been implemented in edge devices such as weeding robots, smart sprayers, and unmanned aerial vehicles (UAVs) for intelligent real-time crop protection~\cite{rai2023applications}. Among the cutting edge YOLO models, YOLOv10-N is at the forefront in terms of lowest latency, slightly higher mAP\textsubscript{50-95}\textsuperscript{val} scores (cf. Table~\ref{tab:yolo-comparison}). 
\begin{table}[ht]
\centering
\caption{Comparison of YOLOv10-N, YOLOv9t, and YOLOv8n based on parameters, average precision (AP), mAP\textsubscript{50-95}\textsuperscript{val}, and inference latency on the COCO detection dataset~\cite{cocodataset} with an image size of 640 pixels, using TensorRT FP16 on a T4 GPU~\cite{wang2024yolov10}}

\label{tab:yolo-comparison}
\begin{tabular}{@{}lccc@{}}
\toprule
\textbf{Model (nano)}   & \textbf{Parameters (M)} & \textbf{Latency (ms)} & \textbf{mAP\textsubscript{50-95}\textsuperscript{val}} \\ \midrule
YOLOv10-N               & 2.3                     & \textbf{1.8}          & \textbf{39.5}          \\
YOLOv9t                 & \textbf{2.0}            & -                     & 38.3                   \\
YOLOv8n                 & 3.2                     & 6.16                  & 37.3                   \\ \bottomrule
\end{tabular}
\end{table}

\subsubsection{Object Detection Metrics}
\label{subsec:metr}
Typically, \textit{precision}, \textit{recall}, \textit{F1 score}, \textit{mAP50}, and \textit{mAP50-95} are popular evaluation metrics for evaluating the performance of object detection models~\cite{Jocher_Ultralytics_YOLO_2023}. \textit{Precision} indicates the proportion of correctly identified positives. In the context of weed detection, high precision means that when the model detects a weed, it is likely correct.
%\begin{equation}
%\text{Precision} = \frac{\text{True Positives}}{\text{True Positives} + \text{False Positives}}
%\label{eq:precision}
%\end{equation}
 \textit{Recall} measures the proportion of actual positives that are correctly identified. In weed detection, it measures how many of the actual weeds present are detected.
%\begin{equation}
%\text{Recall} = \frac{\text{True Positives}}{\text{True Positives} + \text{False Negatives}}
%\label{eq:recall}
%\end{equation}
 To balance \textit{precision} and \textit{recall}, the \textit{F1 score} is calculated. It provides a single metric that balances both precision and recall, reflecting the overall effectiveness of the detection model.
%\begin{equation}
%F1 = 2 \cdot \frac{\text{Precision} \cdot \text{Recall}}{\text{Precision} + \text{Recall}}
%\label{eq:F1}
%\end{equation}

\textit{Intersection over Union (IoU)} is a fundamental metric for object localization in object detection tasks, quantifying the overlap between predicted and ground truth bounding boxes. \textit{Average Precision (AP)} computes the area under the precision-recall curve, providing an overall performance measure of the model, while \textit{Mean Average Precision (mAP)} extends this concept to average precision across all weed classes. \textit{mAP50}~\cite{cocodataset} calculates mAP at an \textit{IoU} threshold of 0.50, while \textit{mAP50-95}~\cite{cocodataset} computes mAP across different IoU thresholds from 0.50 to 0.95. For comprehensive performance evaluation with less localization error, \textit{mAP50-95} is typically preferred~\cite{Jocher_Ultralytics_YOLO_2023}.

\section{Methodological Approach}
\label{sec:meth}
Implementing autonomous weed control by using intelligent agricultural machines (e.g., smart sprayers or robots) is facing a critical trade-off between increasing crop productivity and reducing ecological impact through minimizing use of chemical plant protection products and thus promoting overall agricultural sustainability. However, the limited availability of high-quality and diverse datasets often hinders the training of deep learning (DL) algorithms for autonomous weed detection tasks. Furthermore, due to the high heterogeneity in agricultural fields, available data sets often not reflect the local field conditions properly. 
To address the former issue first, in our approach we generate synthetic data set using a GenAI-based image generation pipeline~\cite{modak2024synthesizing}. 
This section describes our approach by characterizing the dataset used to train the generative AI as well as the synthetic images. 
Subsequently, we elaborate on the training method for the downstream task. We trained compact real-time object detection models (YOLOv8n, YOLOv9t, and YOLOv10-N) using both the original dataset and augmented datasets with synthetic data. The share of augmented data ranged from 10\% to 200\% of the original dataset size. We investigated models pretrained on \textit{COCO} dataset and also when trained \textit{from scratch}. Additionally, our proposed GenAI-based image augmentation approach is compared with traditional image augmentation techniques (copy-paste, mixup, changing hsv color space, and image flipping \&rotating) for each of the abovementioned detection models. We utilized an NVIDIA A100-SXM4-40GB GPU accelerator with 6 GB of memory throughout the entire   Stable Diffusion and YOLO model training and evaluation process.

\subsection{Data set}
\label{subsec:data}
The dataset used in this study consists of a combination of \textit{real-world} data collected in a funded research project (see Acknowledgement~\ref{subsec:ack}) and a \textit{synthetic} dataset generated by text-prompting on a fine-tuned \textit{Stable Diffusion} model~\cite{modak2024synthesizing}. 

\begin{figure}[h]
    \centering
    \begin{subfigure}[b]{0.24\textwidth}
        \centering
        \includegraphics[width=\textwidth]{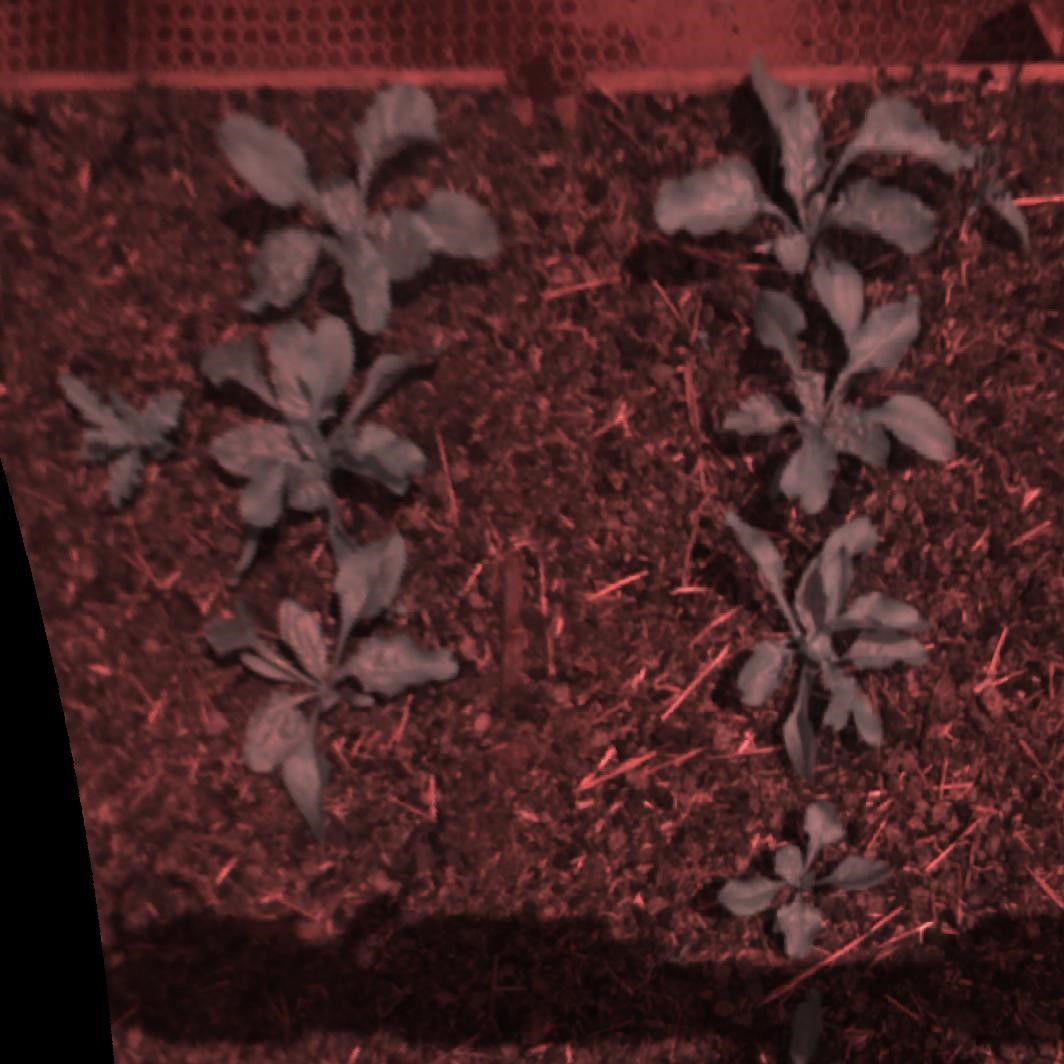}
        \caption{}
        \label{fig:real0}
    \end{subfigure}
    \begin{subfigure}[b]{0.24\textwidth}
        \centering
        \includegraphics[width=\textwidth]{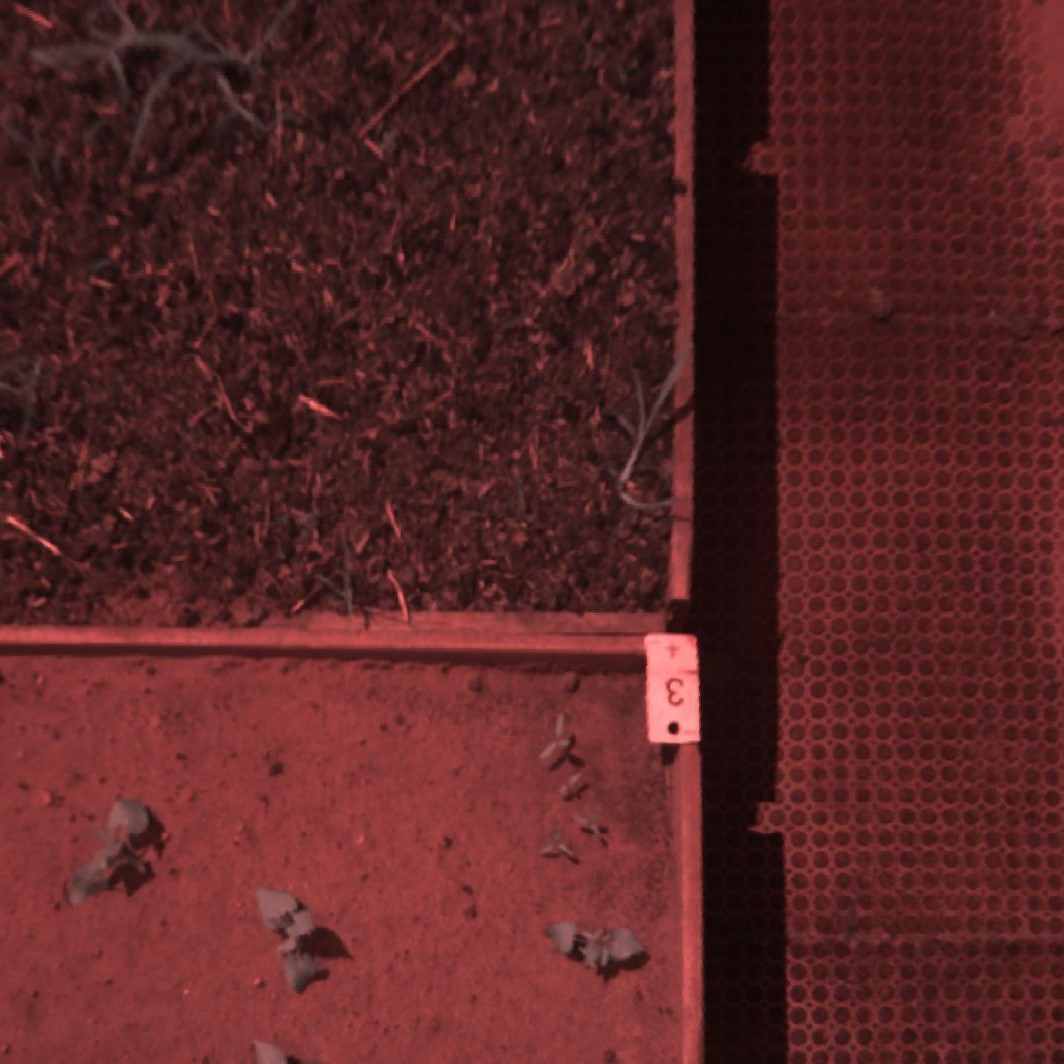}
        \caption{}
        \label{fig:real1}
    \end{subfigure}
    \begin{subfigure}[b]{0.24\textwidth}
        \centering
        \includegraphics[width=\textwidth]{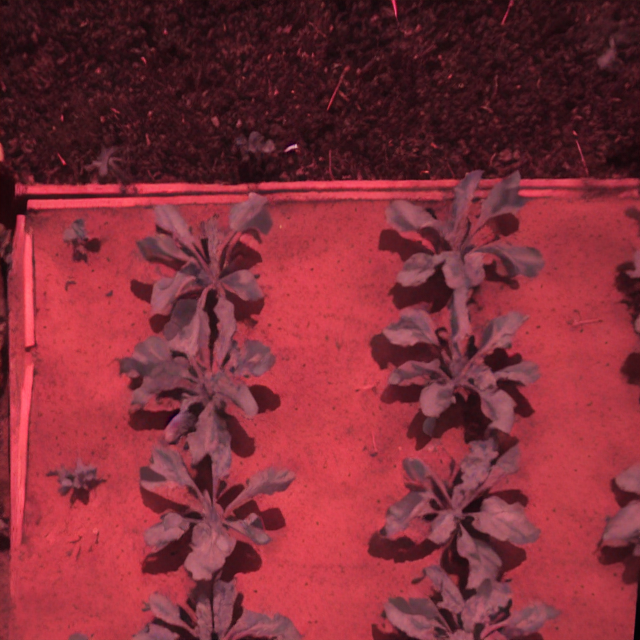}
        \caption{}
        \label{fig:syn0}
    \end{subfigure}
    \begin{subfigure}[b]{0.24\textwidth}
        \centering
        \includegraphics[width=\textwidth]{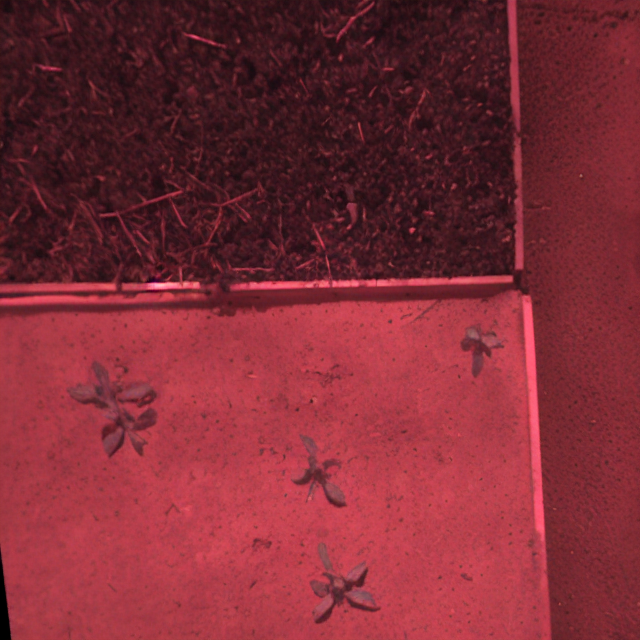}
        \caption{}
        \label{fig:syn1}
    \end{subfigure}
    \caption{Samples of \textit{pseudo-RGB} images: (a) and (b) are from described real-world data sets, while (c) and (d) are synthetic images generated by the image generation pipeline. The dataset comprises main crop Sugar beet and weeds from both real-world and synthetic datasets.}
    \label{fig:sample}
\end{figure}

\paragraph{\textbf{Real-world Images}} The \textit{real-world} dataset was gathered from an experimental site utilizing an advanced \textit{field camera unit (FCU)}. This \textit{FCU} was mounted on a smart sprayer attached to a tractor, operating at a constant speed of \SI{1.5}{\meter\per\second} to ensure consistent image quality. The imaging configuration included an \textit{Effective Focal Length (EFL)} of 6 mm and a 2.3 MP RGB sensor, optimized for high-resolution capture.
The \textit{FCU} featured a dual-band lens filter specifically designed to capture red and near-infrared (NIR) wavelengths. Multiple FCUs were mounted on the sprayer's linkages, maintaining a uniform height of $1.1$ meters above the ground and positioned at a 25-degree off-vertical angle. This arrangement enabled comprehensive coverage and high-quality data acquisition within a controlled outdoor experimental setup. The experimental setup comprised various crops and weeds cultivated under various soil conditions, each distinctly marked on euro pallets for precise identification. This methodology ensured a balanced data set for training robust weed detection models. Post-capture, the raw RED and NIR bands underwent projection correction and subsequently produced pseudo-RGB images. These images were then manually annotated by domain experts with agronomic study background for the object detection task. The resultant data set includes 2074 images, which predominantly feature \textit{Sugar beet} as the primary crop class, along with four weed classes: \textit{Cirsium}, \textit{Convolvulus}, \textit{Fallopia}, and \textit{Echinochloa}. Each image possesses a resolution of 1752 × 1064 pixels, providing detailed visual information necessary for advanced weed detection (cf. Fig ~\ref{fig:sample}).

\begin{figure}[h]
    \centering
    \begin{subfigure}[b]{0.24\textwidth}
        \centering
        \includegraphics[width=\textwidth]{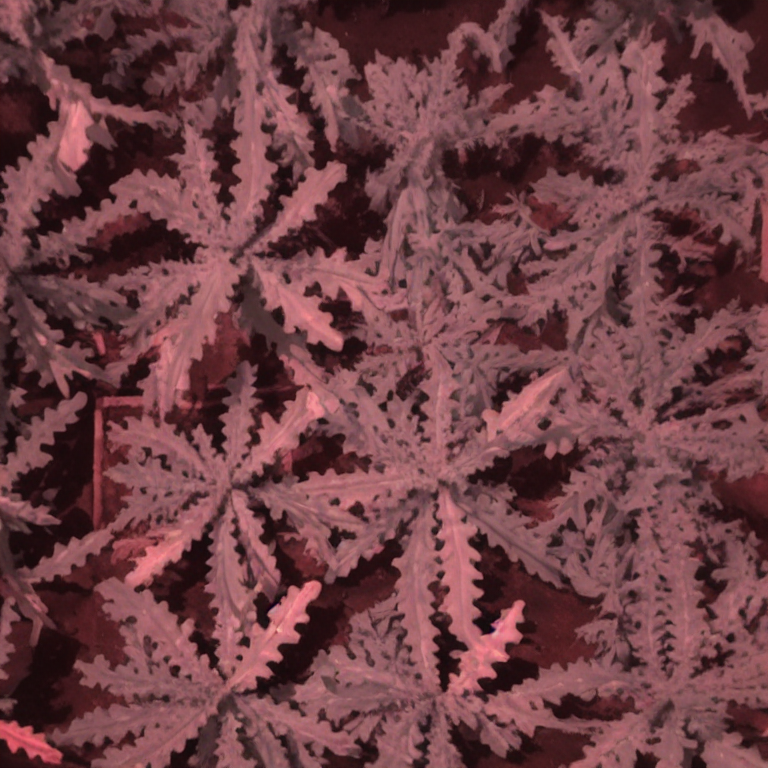}
        \caption{}
        \label{fig:def1}
    \end{subfigure}
    \begin{subfigure}[b]{0.24\textwidth}
        \centering
        \includegraphics[width=\textwidth]{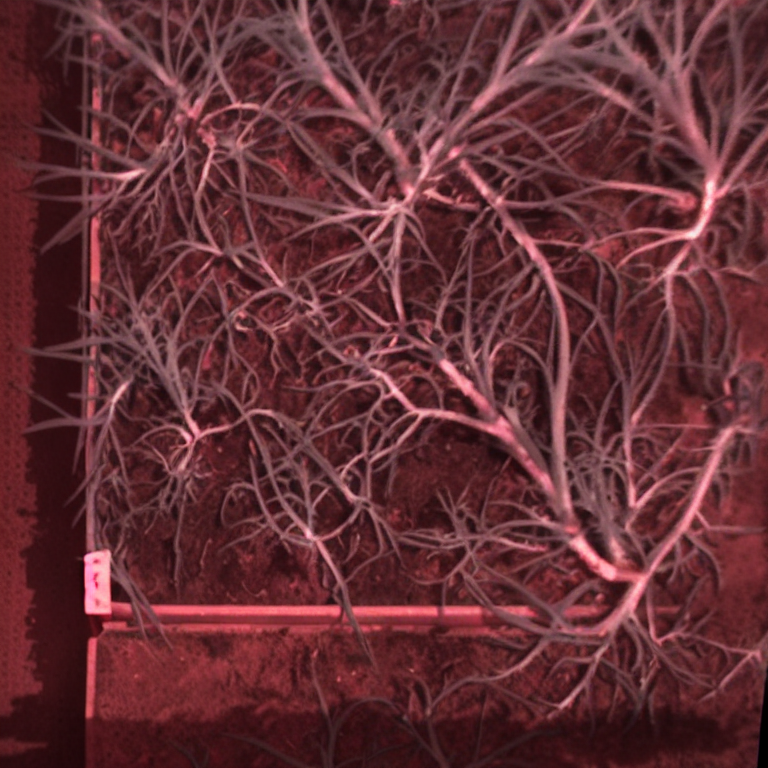}
        \caption{}
        \label{fig:def2}
    \end{subfigure}
    \begin{subfigure}[b]{0.24\textwidth}
        \centering
        \includegraphics[width=\textwidth]{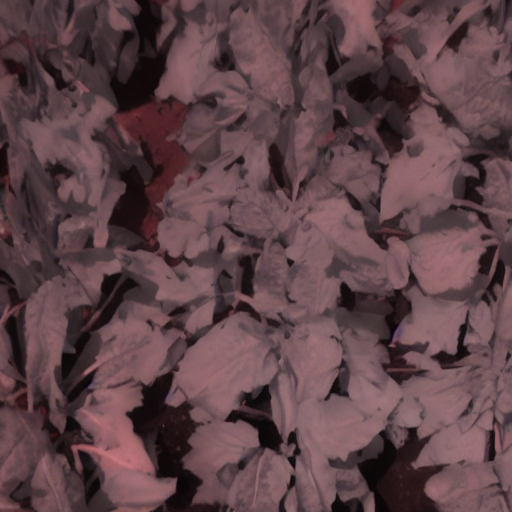}
        \caption{}
        \label{fig:def3}
    \end{subfigure}
    \begin{subfigure}[b]{0.24\textwidth}
        \centering
        \includegraphics[width=\textwidth]{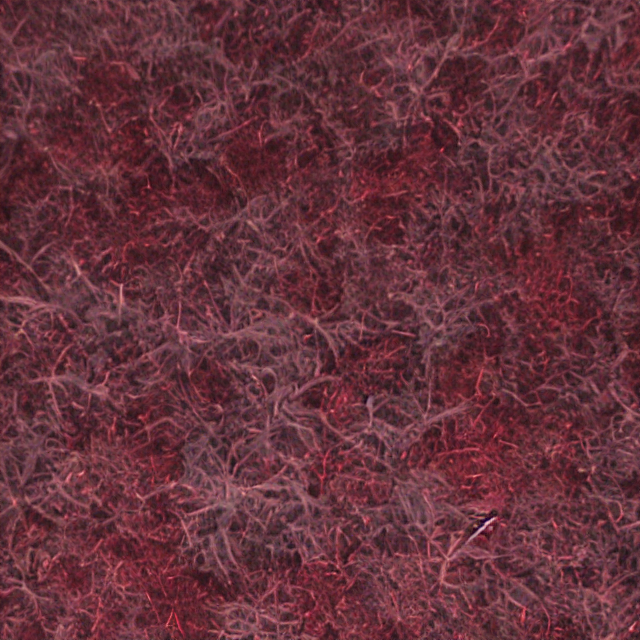}
        \caption{}
        \label{fig:def4}
    \end{subfigure}
    \caption{Examples of problematic synthetic images illustrating various issues: (a) and (b) irregular phenotypic details (e.g., abnormalities in leaf structure); (c) deformed and distorted features; and (d) absence of meaningful content or occurrence of nonsensical patterns.}
    \label{fig:chall}
\end{figure}

\paragraph{\textbf{Synthetic Images}} We adopted a recently proposed, straightforward but robust text-to-image based on Stable Diffusion image-generation pipeline~\cite{modak2024synthesizing} to generate diversified and realistic data (cf. Fig.~\ref{fig:sample}). The pipeline was selected for its ability to efficiently produce high spatial quality (brightness, noisiness, sharpness, complexity), high fidelity, and diverse synthetic images that mimic real-world scenarios. This was measured using a no-reference image quality evaluation metric, namely CLIP-IQA~\cite{wang2023exploring}. This pipeline leverages the foundation model SAM and Stable Diffusion models. SAM was used to convert the annotated real-world images into instance segmentation polygons, followed by mask generation of distinct plant shapes to avoid background nuisance. Based on this,, a \textit{Stable Diffusion $1.5$} model was fine-tuned using the masked plant/weed class and background soils. We used two types of prompts to fine-tune image generation: to address class imbalance, we explicitly used the name of the weed class and soil (plot) in the prompt, such as \textit{ `A photo of Echinochloa, the Sugar beet plot in the background'}; and to introduce data diversity, \textit{ `A photo of random plants and weeds, the Sugar beet plot in the background'} was used, resulting in the generation of approximately 5200 images The distinct characteristics of the real-world pseudo-RGB images are reflected in the synthetic images, creating similar image types. However, following the generation of synthetic images, those exhibiting issues such as deformed features, irregular phenotypic details, or a lack of meaningful content were manually excluded from the dataset (see Fig.~\ref{fig:chall}).

In the initial phase of our study, next to the crop class, we classified four weed species: Cirsium, Convolvulus, Fallopia, and Echinochloa. To enhance practical applicability in weed management, we grouped these species into two categories based on their botanical types: dicotyledons (Cirsium, Convolvulus, Fallopia) and monocotyledons (Echinochloa). This adjustment was made to align with commercially available herbicides that target specific botanical types rather than individual species~\cite{herrera2014novel}. This resulted in a total of three classes: Sugar beet, dicotyledons (Dicot), and monocotyledons (Monocot). Later, we employed a model-guided annotation technique to annotate the synthetic images and trained a YOLOv8x model using manually labeled real-world images for the three class types. The use of the large YOLOv8x model aims to ensure accurate annotation, which can subsequently be extended to an \textit{(inter-)active annotation} process for synthetic images, following the approach of ~\cite{interactive}. We avoid using the large model for the downstream task because our final goal is to deploy the downstream task on edge devices, where nano models are preferred (cf. ~\ref{subsec:yolo}).

\subsection{Experimental settings}
\label{subsec:ex}
\begin{figure}[htbp]
    \centering
    \includegraphics[width=\textwidth]{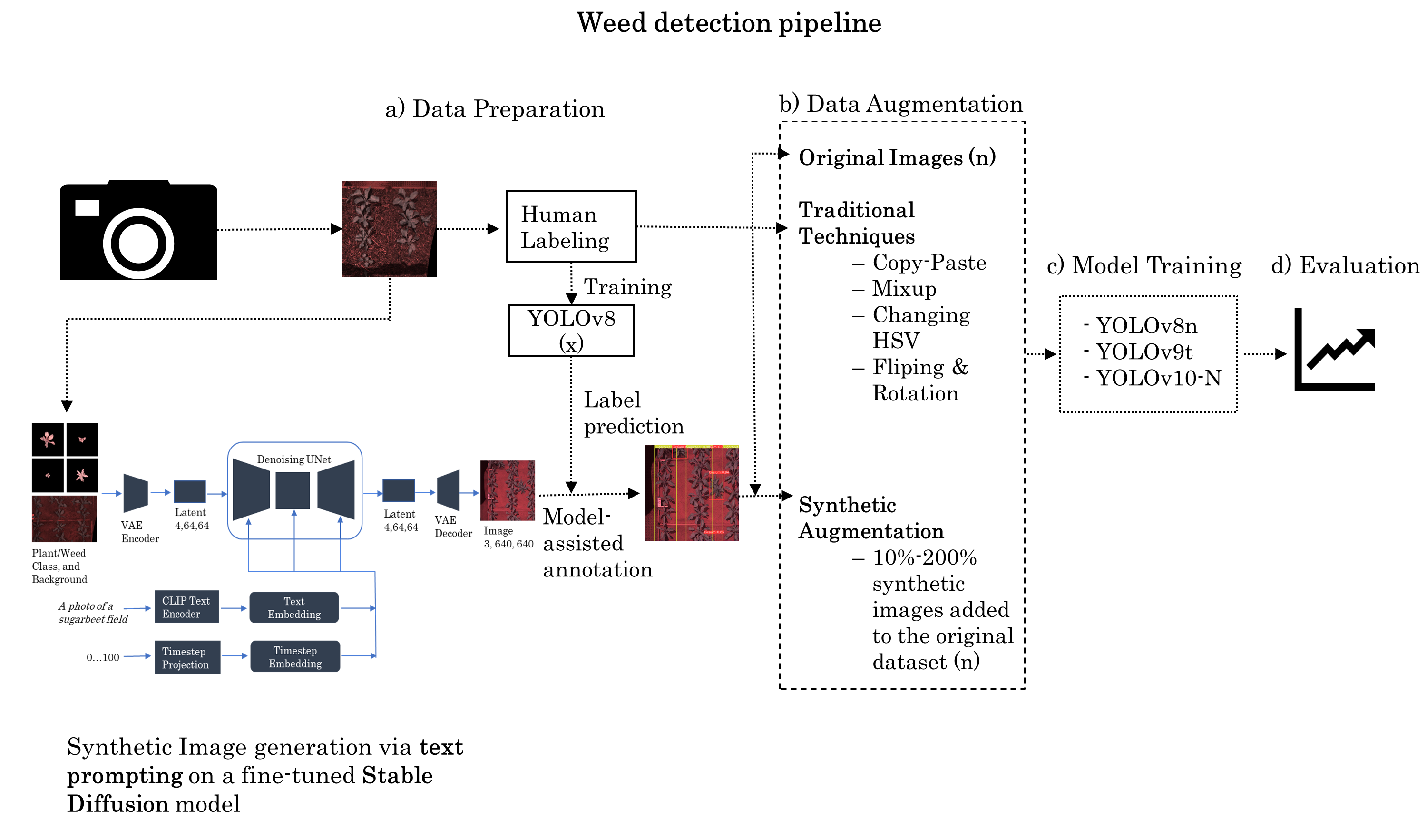}
    \caption{ The weed detection pipeline consists of 4 steps: data preparation (a), data augmentation (b), model training (c), and evaluation (d). Step a) involves data collection via a camera on a smart sprayer and \textit{text-prompting} with a fine-tuned \textit{Stable Diffusion} model. Synthetic images are annotated by a fine-tuned YOLOv8x model using a real-world manually annotated dataset. Step b) applies two types of augmentation: i) Traditional techniques like \textit{copy-paste, mixup, changing HSV, flipping \& rotation}, and ii) augmentation with synthetic images, increasing from 10\% to 200\% in 10\% increments. In step c), three compact YOLO models—YOLOv8n, YOLOv9t, and YOLOv10-N—are used. Finally, step d) evaluates the augmented images.}
    \label{fig:pipe}
\end{figure}

We have trained compact real-time object detectors, in this study YOLO nano models (cf. Sect.~\ref{subsec:yolo}), to compare the augmentation techniques (see Fig.~\ref{fig:pipe}). Two training strategies have been considered: fine-tuned on \textit{(COCO) pretrained weights}, and so-called \textit{training from scratch} for for assessing the reliance on pretrained model weights. The data set was divided into training, validation, and testing sets by 70\%, 15\%, and 15\%, respectively. The hyperparameter configarations for model training are presented in table~\ref{tab:hyp}.

\begin{table}[!ht]
\centering
\caption{Hyperparameter configuration for model training}
\label{tab:hyp}
\begin{tabular}{@{}ll@{}}
\toprule
\textbf{Hyperparameter} & \textbf{Value} \\ \hline
Epochs & 300 \\
Patience &   30  \\
Batch size & 16 \\
Initial learning rate & 0.01 \\
Learning rate schedule & Cosine  \\
\bottomrule
\end{tabular}
\end{table}

During training, \textit{traditional augmentation} techniques were applied by so-called \textit{online augmentation}, which dynamically augments data during training to enhance model generalization without pre-generated augmented datasets. Specifically, we employed four techniques using the Ultralytics library \cite{Jocher_Ultralytics_YOLO_2023}: \textit{copy-paste}, \textit{mixup}, \textit{HSV augmentation}, and \textit{flipping \& rotation}. The \textit{copy-paste} technique enhances diversity by copying random patches from one image and pasting them onto another randomly chosen image. \textit{Mixup} creates composite images by blending multiple images and their labels, promoting generalized feature learning. \textit{HSV augmentation} introduces random changes to Hue, Saturation, and Value, improving the model's robustness to color and lighting variations. \textit{Flipping and rotation} techniques involve horizontal or vertical flipping and rotation, enhancing the model's orientation invariance. Each of these techniques was assigned a probability of \(0.5\), indicating that each image in the training set had a 50\% chance of undergoing that specific augmentation during each epoch of the training process~\cite{Jocher_Ultralytics_YOLO_2023}. This probability of \(0.5\) balances preserving the original dataset with adding enough variability to improve model training. Additionally, to assess these augmentation techniques unbiasedly, we disabled the automated augmentation features of the Ultralytic library.

For \textit{synthetic image augmentation}, artificial training images were sequentially created and randomly added to the training dataset, resulting in \( n = 20 \) augmented datasets, each with a size $T$ of the original training dataset increased by factor $s\in \{10\%, 20\%, \ldots, 200\%\}$ of $T$. 
%\vspace{-8cm}

\section{Evaluation}
\label{sec:eval}
The goal of augmentation is to enhance the robustness and generalizability of object detection models, specifically in our case for weed detection in heterogeneous and diverse agricultural environments of various weed shapes. Therefore, we compare the performance of various traditional augmentation techniques (Copy-paste, HSV, Mixup, Flipping, and rotation) with GenAI-based augmentation using synthetic images generated by a fine-tuned Stable Diffusion model (cf. Section \ref{subsec:img_aug}) and training state-of-the-art compact YOLO models (YOLOv8n, YOLOv9t, YOLOv10-N) for weed detection (see Table \ref{tab:mAp50} \& \ref{table:mAP50-95} ). Performance is assessed using the standard mAP50 and mAP50-95 metrics (cf. Section \ref{subsec:yolo}). In our context, mAP50 measures weed detection accuracy by evaluating predicted bounding boxes against ground truth bounding boxes at an IoU threshold of 0.50, providing an overall indication of model performance independent of task-specifically configured confidence thresholds. On the other hand, mAP50-95 averages average precision scores over IoU thresholds from 0.50 to 0.95, offering insights into the model's ability to detect weeds across different requirements for accuracy of bounding box matching. This is crucial for tasks such as intelligent weed control, where precision in identifying weeds of various sizes, shapes, environmental conditions or occlusion/overlapping is essential. A high mAP50-95 indicates higher accuracy across all weed types, without bias toward easier-to-detect weeds. Conversely, a model with high mAP50 but low mAP50-95 might perform well in detecting larger, more obvious weeds but struggle with smaller, harder-to-detect ones~\cite{Uncut}. Our experiments show that all augmentation techniques consistently improved the mAP50 (cf. Table~\ref{tab:mAp50}) and mAP50-95 scores (cf. Table~\ref{table:mAP50-95}) on the \textit{test set} across all investigated versions of YOLO.

Considering the mAP50 metric (see Table~\ref{tab:mAp50}) first, the YOLOv8n(COCO) model, fine-tuned on \textit{COCO-pretrained weights}, exhibited an mAP50 increase of up to 2\% when augmented with the \textit{Original + Synthetic (50\%)} and \textit{Original + Synthetic (200\%)} datasets, resulting in a value of $\approx$ 0.89. Similar improvements were observed in both the YOLOv9t(COCO) and YOLOv10-N(COCO) models. Specifically, the YOLOv9t(COCO) model demonstrated an increase in mAP50 of up to 3\% when mixup-based augmentation and the \textit{Original + Synthetic (130\%)} data set were applied, resulting in an mAP50 of 0.90. The YOLOv10-N(COCO) model experienced a 4\% improvement with the \textit{Original + Synthetic (80\%)} dataset, reaching a peak mAP50 of 0.86. \newline When \textit{training from scratch}, the YOLOv8n (scratch) model showed a significant increase in mAP50 of 20\%, increasing from 0.608 to 0.82 with the \textit{ original + synthetic (40\%) dataset }. 
The YOLOv9t (Scratch) model achieved a remarkable 27\% enhancement in mAP50 (mAP50=0.87) 
over the baseline (mAP50=0.608) when utilizing the \textit{Original + Synthetic (80\%)} and \textit{Original + Synthetic (100\%)} datasets. Furthermore, the YOLOv10-N(Scratch) model, trained from scratch, demonstrated a substantial improvement in mAP50 of 30\% with the \textit{Original + Synthetic (190\%)} data set, culminating in a mAP50 score of 0.77.

\begin{center}

\scriptsize
\begin{longtable}{p{2.8cm} *{6}{>{\centering\arraybackslash}p{1.45cm}}}

\caption{Comparison of traditional augmentation techniques (Copy-Paste, HSV, Mixup, Flipping \& Rotation) and synthetic image augmentation using the Stable Diffusion model on three compact YOLO models (YOLOv8n, YOLOV9t, YOLOv10-N), either trained with COCO-pretrained weights (\textbf{YOLO (COCO)}) or `from scratch' (\textbf{YOLO (Scratch)}). Results are presented in terms of \textbf{mAP50} metrics; the best results are highlighted in bold.}
\label{tab:mAp50}\\
\toprule
Augmentation & \textbf{YOLOv8n (COCO)} & \textbf{YOLOv8n (Scratch)} & \textbf{YOLOv9t (COCO)} & \textbf{YOLOv9t (Scratch)} & \textbf{YOLOv10-N (COCO)} & \textbf{YOLOv10-N (Scratch)} \\
\midrule
\endfirsthead

\midrule
Augmentation & \textbf{YOLOv8n (COCO)} & \textbf{YOLOv8n (Scratch)} & \textbf{YOLOv9t (COCO)} & \textbf{YOLOv9t (Scratch)} & \textbf{YOLOv10-N (COCO)} & \textbf{YOLOv10-N (Scratch)} \\
\midrule
\endhead

\midrule
\endfoot

\bottomrule
\endlastfoot

No Augmentation & 0.872 & 0.608 & 0.874 & 0.608 & 0.817 & 0.469 \\
Copy-paste & 0.882 & 0.707 & 0.894 & 0.749 & 0.801 & 0.657 \\
HSV & 0.874 & 0.671 & 0.889 & 0.725 & 0.782 & 0.545 \\
Mix & 0.882 & 0.673 & \textbf{0.896} & 0.836 & 0.815 & 0.684 \\
Flip and rot. & 0.881 & 0.701 & 0.883 & 0.730 & 0.820 & 0.636 \\
Orig. + Synth. (10\%) & 0.884 & 0.801 & 0.884 & 0.749 & 0.808 & 0.535 \\
Orig. + Synth. (20\%) & 0.879 & 0.783 & 0.882 & 0.717 & 0.842 & 0.550 \\
Orig. + Synth. (30\%) & 0.876 & 0.773 & 0.885 & 0.785 & 0.806 & 0.513 \\
Orig. + Synth. (40\%) & 0.860 & \textbf{0.821} & 0.866 & 0.714 & 0.816 & 0.551 \\
Orig. + Synth. (50\%) & \textbf{0.888} & 0.737 & 0.873 & 0.840 & 0.834 & 0.646 \\
Orig. + Synth. (60\%) & 0.857 & 0.744 & 0.878 & 0.749 & 0.817 & 0.633 \\
Orig. + Synth. (70\%) & 0.885 & 0.718 & 0.873 & 0.877 & 0.805 & 0.467 \\
Orig. + Synth. (80\%) & 0.878 & 0.756 & 0.884 & 0.849 & \textbf{0.860} & 0.632 \\
Orig. + Synth. (90\%) & 0.867 & 0.787 & 0.878 & 0.808 & 0.851 & 0.672 \\
Orig. + Synth. (100\%) & 0.873 & 0.734 & 0.884 & \textbf{0.878} & 0.828 & 0.736 \\
Orig. + Synth. (110\%) & 0.884 & 0.750 & 0.883 & 0.857 & 0.823 & 0.628 \\
Orig. + Synth. (120\%) & 0.871 & 0.783 & 0.892 & 0.612 & 0.845 & 0.584 \\
Orig. + Synth. (130\%) & 0.854 & 0.803 & \textbf{0.896} & 0.796 & 0.842 & 0.633 \\
Orig. + Synth. (140\%) & 0.871 & 0.792 & 0.881 & 0.755 & 0.854 & 0.704 \\
Orig. + Synth. (150\%) & 0.859 & 0.787 & 0.882 & 0.857 & 0.825 & 0.659 \\
Orig. + Synth. (160\%) & 0.873 & 0.767 & 0.889 & 0.830 & 0.810 & 0.691 \\
Orig. + Synth. (170\%) & 0.867 & 0.775 & 0.874 & 0.872 & 0.836 & 0.655 \\
Orig. + Synth. (180\%) & 0.870 & 0.780 & 0.873 & 0.865 & 0.803 & 0.716 \\
Orig. + Synth. (190\%) & 0.870 & 0.694 & 0.879 & 0.858 & 0.816 & \textbf{0.770} \\
Orig. + Synth. (200\%) & \textbf{0.888} & 0.757 & 0.885 & 0.867 & 0.809 & 0.689 \\
\end{longtable}
\end{center}

\begin{center}
\scriptsize
\begin{longtable}{p{2.8cm} *{6}{>{\centering\arraybackslash}p{1.45cm}}}
\caption{Comparison of traditional augmentation techniques (Copy-Paste, HSV, Mixup, Flipping \& Rotation) and synthetic image augmentation using the Stable Diffusion model on three compact YOLO models (YOLOv8n, YOLOV9t, YOLOv10-N), either trained with COCO-pretrained weights (\textbf{YOLO (COCO)}) or `from scratch' (\textbf{YOLO (Scratch)}). Results are presented in terms of  \textbf{mAP50-95} metric; the best results are highlighted in bold.}
\label{table:mAP50-95}\\
\toprule
\textbf{Augmentation} & \textbf{YOLOv8n (COCO)} & \textbf{YOLOv8n (Scratch)} & \textbf{YOLOv9t (COCO)} & \textbf{YOLOv9t (Scratch)} & \textbf{YOLOv10-N (COCO)} & \textbf{YOLOv10-N (Scratch)} \\
\midrule
\endfirsthead

\midrule
\textbf{Augmentation} & \textbf{YOLOv8n (COCO)} & \textbf{YOLOv8n (Scratch)} & \textbf{YOLOv9t (COCO)} & \textbf{YOLOv9t (Scratch)} & \textbf{YOLOv10-N (COCO)} & \textbf{YOLOv10-N (Scratch)} \\
\midrule
\endhead

\midrule
\endfoot

\bottomrule
\endlastfoot

No Augmentation & 0.679 & 0.384 & 0.666 & 0.384 & 0.603 & 0.326 \\
Copy-paste & 0.680 & 0.470 & 0.670 & 0.530 & 0.593 & 0.438 \\
HSV & 0.651 & 0.435 & 0.671 & 0.508 & 0.571 & 0.350 \\
Mix & 0.676 & 0.435 & 0.660 & 0.607 & 0.608 & 0.461 \\
Flip and rot. & 0.694 & 0.482 & 0.638 & 0.507 & 0.575 & 0.396 \\
Orig. + Synth. (10\%) & \textbf{0.720} & \textbf{0.637} & 0.711 & 0.502 & 0.620 & 0.362 \\
Orig. + Synth. (20\%) & 0.712 & 0.590 & 0.708 & 0.507 & 0.640 & 0.376 \\
Orig. + Synth. (30\%) & 0.710 & 0.594 & 0.703 & 0.581 & 0.628 & 0.326 \\
Orig. + Synth. (40\%) & 0.708 & 0.607 & 0.706 & 0.471 & 0.646 & 0.362 \\
Orig. + Synth. (50\%) & 0.700 & 0.521 & 0.697 & 0.650 & 0.658 & 0.497 \\
Orig. + Synth. (60\%) & 0.683 & 0.512 & 0.720 & 0.526 &\textbf{ 0.676} & 0.431 \\
Orig. + Synth. (70\%) & 0.704 & 0.531 & 0.692 & 0.697 & 0.660 & 0.315 \\
Orig. + Synth. (80\%) & 0.716 & 0.564 & \textbf{0.724} & 0.641 & 0.660 & 0.411 \\
Orig. + Synth. (90\%) & 0.691 & 0.551 & 0.703 & 0.621 & 0.652 & 0.457 \\
Orig. + Synth. (100\%) & 0.710 & 0.544 & 0.689 & \textbf{0.709} & 0.640 & 0.537 \\
Orig. + Synth. (110\%) & 0.702 & 0.558 & 0.693 & 0.703 & 0.649 & 0.465 \\
Orig. + Synth. (120\%) & 0.694 & 0.576 & 0.693 & 0.429 & 0.659 & 0.426 \\
Orig. + Synth. (130\%) & 0.695 & 0.628 & 0.714 & 0.574 & 0.666 & 0.420 \\
Orig. + Synth. (140\%) & 0.711 & 0.506 & 0.698 & 0.593 & 0.654 & \textbf{0.554} \\
Orig. + Synth. (150\%) & 0.697 & 0.604 & 0.694 & 0.691 & 0.636 & 0.462 \\
Orig. + Synth. (160\%) & 0.694 & 0.559 & 0.707 & 0.638 & 0.640 & 0.529 \\
Orig. + Synth. (170\%) & 0.686 & 0.615 & 0.697 & 0.703 & 0.663 & 0.455 \\
Orig. + Synth. (180\%) & 0.709 & 0.599 & 0.691 & 0.694 & 0.617 & 0.552 \\
Orig. + Synth. (190\%) & 0.688 & 0.428 & 0.709 & 0.679 & 0.646 & 0.475 \\
Orig. + Synth. (200\%) & 0.703 & 0.594 & 0.705 & 0.687 & 0.632 & 0.513 \\
\end{longtable}
\end{center}

Moving the focus to the mAP50-95 metric (see Table \ref{table:mAP50-95}), in becomes apparent that the YOLOv8n (COCO) model -- fine-tuned on \textit{COCO-pretrained weights} -- evaluated at approximately 0.72 when employing the \textit{Original + Synthetic (10\%)} dataset, reflecting an augmentation from a baseline score of 0.679. The YOLOv9t (COCO) model demonstrated an elevation in mAP50-95 to 0.724 with the \textit{Original + Synthetic (80\%)} dataset, up from 0.666 without augmentation. The YOLOv10-N (COCO) model achieved a maximum mAP50-95 of 0.676 with the \textit{Original + Synthetic (60\%)} dataset, compared to a baseline of 0.603. Among the traditional augmentation techniques, the copy-paste based augmentation slightly improved the mAP50-95 scores except for the YOLOv10-N (COCO). Moreover, the application of HSV-based augmentation negatively affected the performance of both the YOLOv8n (COCO) and YOLOV10-N (COCO) models.
For models trained \textit{from scratch}, the YOLOv8n (Scratch) exhibited a significant increase in mAP50-95 scores, rising to 0.637 with the \textit{Original + Synthetic (10\%)} dataset from a baseline of 0.384 without augmentation. The YOLOv9t (Scratch) model showcased a remarkable mAP50-95 increment to 0.709 with the \textit{Original + Synthetic (100\%)} dataset, as opposed to 0.384 without augmentation. The YOLOv10-N (Scratch) model presented a notable improvement in mAP50-95 to 0.554 with the \textit{Original + Synthetic (140\%)} dataset, in comparison to a baseline score of 0.326.

\section{Discussion} 
\label{sec:dis}
The reported results indicate that incorporating GenAI generated synthetic images positively influence model performance and accuracy (measured by mAP50-95 score). Nevertheless, conventional image augmentation techniques have also been found to enhance mAP50 scores, demonstrating their effectiveness for model improvement (see Table \ref{tab:mAp50}). Advanced techniques such as copy-paste and mixup consistently outperform simpler methods such as HSV adjustment, flip and rotation, achieving results comparable to synthetic image augmentation when fine-tuning COCO-pretrained weights. This suggests that until model robustness becomes a primary challenge and computational resources are limited, traditional yet advanced image augmentation techniques (such as copy-paste and mix-up) can be advantageous for model training. However, their efficacy might decrease in complex scenarios that entail highly varying domains due to heterogeneous environmental conditions (lighting, soil color, weather condition etc.) which in turn would require both diverse and still realistic data variations.

In contrast, synthetic data enriches the dataset with entirely new, unseen examples beyond real-world data constraints, offering essential diversity in object appearances, backgrounds, and contexts. This diversity enables models to learn robust features and generalize effectively to unseen data distributions. Despite their benefits, synthetic data augmentation techniques pose practical challenges, including high computational resource demands and the need for careful quality control by the domain experts. Traditional augmentation during training may require fewer computational resources but still faces significant challenges in improving model robustness. 
So far, in this study we investigated how augmentation of training data leads to improved model performance. However, in view of the still necessary manual step of annotating high quality training data, the potential of a GenAI-based augmentation approach for effectively reducing the amount of needed human-annotated `real' data is currently investigated. Finally, the study highlights the significance of including synthetic data in training pipelines to enhance the accuracy of weed detection systems.

\section{Conclusion}
\label{sec:con}

This paper demonstrates the promising potential of GenAI-generated synthetic data in enhancing model training for highly-precise weed detection. By augmenting the data sets used to train compact object detection models, specifically YOLO versions v8n, v9t, and v10-N, we achieved notable improvements in model performance. Synthetic data augmentation proved effective in diversifying datasets with novel examples, thereby enhancing model generalization across various data distributions. However, the approach highlights practical challenges, including high computational resource demands and the necessity for rigorous quality control. Although traditional augmentation methods such as copy-paste and mix-up remain valuable, especially under resource constraints, their efficacy reduces in complex scenarios requiring diverse and realistic data representations, such as weed detection in Sugar beet image.

Future research should focus on developing hybrid augmentation strategies that combine the strengths of both traditional and generative AI-based techniques for more effective model training. In addition, traditional offline image augmentation methods will be explored, with progressive addition of data, similar to the synthetic image augmentation approach as investigated in this work. Moreover, optimizing generative AI techniques for computational efficiency will be crucial for their broader adoption in resource-limited embedded devices such as weeding robots. 

To expand the scope and diversity of this study, incorporating multiple datasets encompassing diverse soil types and plant/weed varieties could further enhance the utility of generative models. Our findings indicate that GenAI-based augmentation improves object detection capabilities and improves model accuracy. For further analysis, integrating explainable AI techniques such as Gradient-weighted Class Activation Mapping (Grad-CAM)~\cite{selvaraju2017grad} could provide insights into how the model identifies and localizes small and complex weed scenes, as explored in a study~\cite{xu2024weedsnet}, and how the targeted creation of synthetic images might enhance the model training to specifically support the reliability of localization and classification.

%\clearpage  % TODO REVIEW/FINAL: This \clearpage needs to be removed from both review and camera-ready versions.

\subsubsection*{Acknowledgements.}
\label{subsec:ack}
This research was conducted within the scope of the project ``Hochleistungssensorik für smarte Pflanzenschutzbehandlung (HoPla)'' (FKZ 13N16327), and is supported by the Federal Ministry of Education and Research (BMBF) and VDI Technology Center on the basis of a decision by the German Bundestag.
\bibliographystyle{splncs04}
\bibliography{ref}
\end{document}